\documentclass[10pt,twocolumn,letterpaper]{article}

\usepackage{cvpr}
\usepackage{times}
\usepackage{epsfig}
\usepackage{graphicx}
\usepackage{subfigure}
\usepackage{multirow}
\usepackage{amsmath}
\usepackage{amssymb}
\usepackage{bm}
\usepackage{xcolor}
\usepackage{booktabs}
\usepackage[pagebackref=true,breaklinks=true,letterpaper=true,colorlinks,bookmarks=false]{hyperref}

\cvprfinalcopy % *** Uncomment this line for the final submission

\definecolor{battleshipgrey}{rgb}{0.52, 0.52, 0.51}
\definecolor{capri}{rgb}{0.0, 0.75, 1.0}

\ifcvprfinal\fi
\begin{document}

%%%%%%%%% TITLE
%\title{Person of Attention}
\title{Harmonious Attention Network for Person Re-Identification}
\author{Wei Li$^1$ 
	\quad \quad  \quad \quad  \quad Xiatian Zhu$^2$ 
	 \quad \quad \quad \quad \quad  Shaogang Gong$^1$ \\
Queen Mary University of London$^1$ \quad \quad \quad \quad %\\
% London E1 4NS, United Kingdom 
Vision Semantics Ltd.$^2$\\
{\tt\small \{wei.li, s.gong\}@qmul.ac.uk} 
\quad \quad 
\tt\small eddy@visionsemantics.com
}

\maketitle
%\thispagestyle{empty}

%%%%%%%%% ABSTRACT
\begin{abstract}
	
Existing person re-identification (re-id) methods 
either assume the availability of well-aligned person bounding box images as model input
or rely on constrained attention selection mechanisms to calibrate misaligned images.
They are therefore sub-optimal for re-id matching in arbitrarily
aligned person images potentially with large human pose variations and 
unconstrained auto-detection errors.
In this work, we show the
advantages of jointly learning attention selection and feature representation
in a Convolutional Neural Network (CNN) by 
maximising the complementary information of different levels of visual
attention subject
to re-id discriminative learning constraints. 
Specifically, we
formulate a novel Harmonious Attention CNN (HA-CNN) model for 
joint learning of soft pixel attention and hard regional attention
along with simultaneous optimisation of feature representations, 
dedicated to optimise person re-id in uncontrolled (misaligned) images.
Extensive comparative evaluations validate the superiority of this new
HA-CNN model for person re-id over a wide variety of
state-of-the-art methods on three large-scale benchmarks
including CUHK03, Market-1501, and DukeMTMC-ReID. 	

\end{abstract}

\section{Introduction}
\begin{figure}[!h]
	\centering
	\includegraphics[width=0.8 \linewidth]{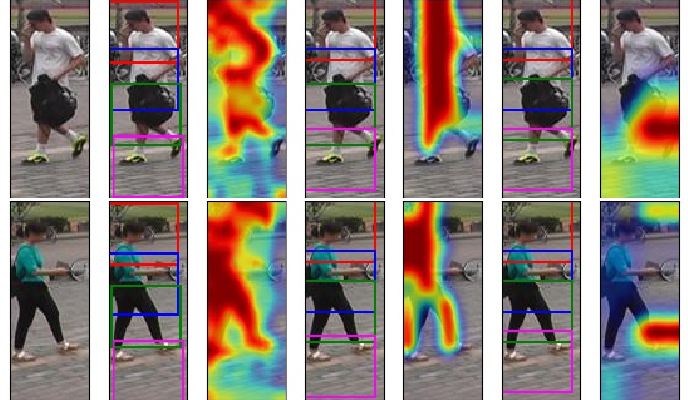}
	\caption{Examples of attention selection in auto-detected person bounding
          boxes used for person re-id matching.
%		{\bf {\color{red} TODO: 
%				(1) Show the misalignment problem in pairwise images.
%				(2) Show the hard and soft attention by our model.}}
	}
	\label{fig:reid}
	\vspace{-0.3cm}
\end{figure}

Person re-identification (re-id) aims to search people across non-overlapping surveillance camera views deployed at different locations by matching person images.
In practical re-id scenarios, 
person images are typically automatically detected for scaling up to
large visual data \cite{zheng2015scalable,li2014deepreid}.
Auto-detected person bounding boxes are typically not optimised for re-id 
due to misalignment with background clutter, occlusion, missing body parts
(Fig. \ref{fig:reid}). 
Additionally, people ({uncooperative)} % in the public spaces
are often captured in various poses across open space and time.
These give rise to the notorious image matching {\em misalignment} challenge
in cross-view re-id \cite{gong2014person}.
% unknown changes in illumination, occlusion, and background clutter is normal \cite{gong2014person}
%
There is consequently an inevitable need for {\em attention selection} 
within arbitrarily-aligned bounding boxes 
as an integral part of model learning for re-id.

%%%%%%%%% BODY TEXT

There are a few attempts in the literature for solving
the problem of re-id attention selection within person bounding boxes.
% hand-crafted feature based
One common strategy is local patch calibration and saliency weighting
in pairwise image matching
\cite{SalienceReId_CVPR13,shen2015person,zheng2015partial,hanxiao2014GTS}.
However, these methods rely on hand-crafted features
without deep learning jointly more expressive feature representations
and matching metric holistically (end-to-end).
% % deep feature based
A small number of attention deep learning models for re-id have been recently developed
for reducing the negative effect 
from poor detection and human pose change
%by combining the regional attention selection and person re-id into a single network
\cite{li2017learning,zhao2017deeply,su2017pose}.  
%lan2017deep
Nevertheless, these deep methods implicitly assume the availability of large labelled training data
by simply adopting existing deep architectures with high complexity in model design.
Additionally, they often consider only coarse region-level attention whilst
ignoring the fine-grained pixel-level saliency.
Hence, these techniques are ineffective when only a small set of labelled data is
available for model training whilst also facing noisy person images of
arbitrary misalignment and background clutter. 

In this work, we consider the problem of jointly deep learning attention selection
and feature representation for optimising person re-id in a more
lightweight (with less parameters) network architecture.
The {\bf contributions} of this work are:
{\bf (I)} 
We formulate a novel idea of jointly learning 
%Harmonious Attention Convolutional Neural Network (HA-CNN) for 
multi-granularity attention selection and feature representation 
for optimising person re-id
in deep learning. 
To our knowledge, this is the first attempt of jointly deep learning 
multiple complementary attention for solving the person re-id problem.
{\bf (II)} 
We propose a {\em Harmonious Attention Convolutional Neural Network} (HA-CNN)
to simultaneously learn hard region-level and soft pixel-level
attention within arbitrary person bounding boxes along with re-id feature representations 
for maximising the correlated complementary information
between attention selection and feature discrimination.  
This is achieved by devising a lightweight Harmonious Attention module
capable of efficiently and effectively learning different types of attention
from the shared re-id feature representation in a multi-task and end-to-end learning fashion.
{\bf (III)}
We introduce a cross-attention interaction learning scheme
for further enhancing the compatibility between attention 
selection and feature representation given re-id discriminative constraints.
Extensive comparative evaluations demonstrate the superiority of the
proposed HA-CNN model over a wide range of state-of-the-art
re-id models on three large benchmarks
CUHK03~\cite{li2014deepreid}, Market-1501~\cite{zheng2015scalable},
and DukeMTMC-ReID \cite{zheng2017unlabeled}.

%-------------------------------------------------------------------------
\section{Related Work}

Most existing person re-id methods
focus on supervised learning of identity-discriminative information,
including ranking by pairwise constraints
\cite{Anton_2015_CoRR,wang2016pami},
discriminative distance metric learning~\cite{KISSME_CVPR12,
	PRD_PAMI13,xiong2014person,liao2015person,zhang2016learning,chen2017person},
and deep learning~\cite{qian2017multi,
	li2014deepreid,chen2016multi,xiao2016learning,wangjoint,li2017person}.
These methods assume that person images are well aligned,
which is largely invalid given imperfect detection bounding boxes of
changing human poses.
To overcome this limitation, 
attention selection techniques have been developed
for improving re-id by %identifying part correspondence, 
localised patch matching~\cite{shen2015person,zheng2015partial} and
saliency weighting~\cite{hanxiao2014GTS,SalienceReId_CVPR13}.
These are inherently unsuitable by design to 
cope with poorly aligned person images, due to their
stringent requirement of tight bounding boxes around the whole person
and high sensitivity of the hand-crafted features.

Recently, a few attention deep learning methods 
have been proposed to handle the matching misalignment challenge
in re-id
\cite{li2017learning,zhao2017deeply,su2017pose, lan2017deep}. 
The common strategy of these methods is to % establish a CNN architecture 
incorporate a regional attention selection sub-network into 
a deep re-id model.
For example, Su et al. \cite{su2017pose} integrate a separately trained pose detection model 
(from additional labelled pose ground-truth) into a part-based re-id model.
Li et al. \cite{li2017learning} design an end-to-end trainable part-aligning CNN network for 
locating latent discriminative regions (i.e. hard attention) and subsequently
extract and exploit these regional features for performing re-id.
Zhao et al. \cite{zhao2017deeply} exploit
the Spatial Transformer Network \cite{jaderberg2015spatial}
as the hard attention model for searching re-id discriminative parts 
given a pre-defined spatial constraint. 
%{\color{red} Xu et al. \cite{lan2017deep} formulate a reinforcement attention 
%model for person region selection under identity discriminative constraints.}
%
However, these models fail to consider the noisy information within
selected regions at the pixel level, i.e. no soft attention modelling,
which can be important.
While soft attention modelling for re-id is considered in
\cite{liu2017hydraplus}, this model assumes tight person boxes thus
less suitable for poor detections. 

The proposed HA-CNN model is designed particularly to
address the weaknesses of existing deep methods as above
by formulating a joint learning scheme for modelling both soft and hard attention
in a single re-id deep model.
This is the first attempt of modelling multi-level correlated attention in deep learning
for person re-id to our knowledge.
In addition, % to simultaneous learning of attention selection and feature representations,
we introduce cross-attention interaction learning for enhancing the complementary
effect between different levels of attention subject to re-id
discriminative constraints. This is impossible to do for existing
methods due to their inherent single level attention modelling. We
show the benefits of joint modelling multi-level attention in person
re-id in our experiments. Moreover, we also design an efficient attention CNN architecture for 
improving the model deployment scalability,
an under-studied but practically important issue for re-id.

\begin{figure*} [h]%[h!]
\centering
\includegraphics[width=0.9 \linewidth]{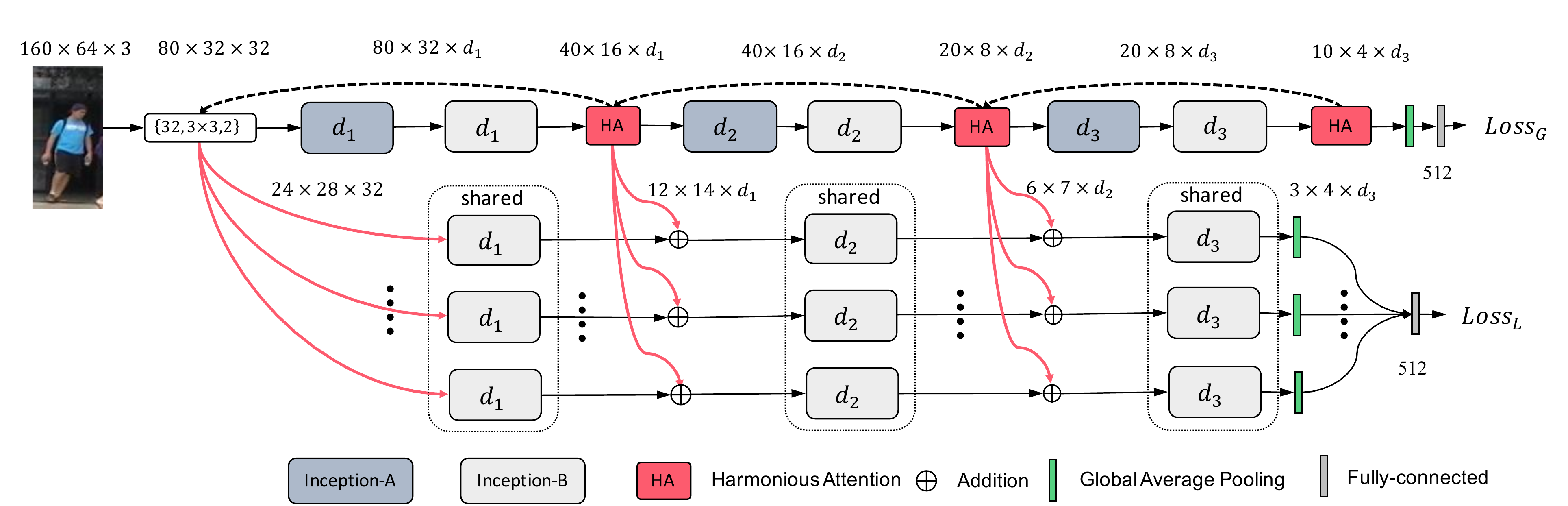}
\caption{ %\footnotesize
	The Harmonious Attention Convoluntional Neural Network.
	The symbol $d_l$ ($l\in\{1,2,3\}$) denotes the number of convolutional filter in the corresponding Inception unit at the $l$-th block.
	}
	\label{fig:pipline}
	\vspace{-0.5cm}
\end{figure*}

\section{Harmonious Attention Network}

Given $n$ training bounding box images $\mathcal{I} = \{ \bm{I}_i \}_{i=1}^n$  
from $n_\text{id}$ distinct people captured by non-overlapping camera views together with the corresponding identity labels as $\mathcal{Y} = \{y_i\}_{i=1}^n$
(where $y_i \in [1,\cdots, n_\text{id}]$), 
we aim to learn a deep feature representation model % both local and global feature representation models 
optimal for person re-id matching under significant viewing condition variations. % across locations. 
To this end, we formulate a {\em Harmonious Attention Convolutional Neural Network} (HA-CNN) 
that aims to concurrently learn a set of harmonious attention, global and local feature representations for maximising their complementary benefit and compatibility 
in terms of both discrimination power and architecture simplicity.
Typically, person parts location information is not provided in person
re-id image annotation (i.e. only weakly labelled without fine-grained).
Therefore, the attention model learning is {\em weakly supervised}
in the context of optimising re-id performance.
Unlike most existing works that simply adopting a standard deep CNN
network typically with a large number of model parameters (likely
overfit given small size labelled data) and high
computational cost in model deployment
\cite{krizhevsky2012imagenet,simonyan2014very,szegedy2015going,he2016deep}, 
we design a {\em lightweight} (less parameters) yet deep (maintaining strong
discriminative power) CNN architecture by
devising a {\em holistic} attention mechanism for
locating the most discriminative pixels and regions in order to identify
optimal visual patterns for re-id. We avoid simply
stacking many CNN layers to gain model depth. 

This is particularly critical for re-id
where the label data is often sparse 
(large models are more likely to overfit in training) 
and the deployment efficiency is very important 
(slow feature extraction is not scalable to large surveillance video data).

\vspace{0.1cm}
\noindent {\bf HA-CNN Overview}
We consider a multi-branch network architecture for our purpose. The
overall objective of this multi-branch scheme and the overall
architecture composition is to minimise the model complexity therefore
reduce the network parameter size whilst maintaining the optimal
network depth. The overall design of our HA-CNN architecture is shown
in Fig. \ref{fig:pipline}.
This HA-CNN model contains two branches:
{\bf (1)} One {\em local branch} (consisting of $T$ streams of an identical structure):
Each stream aims to learn the most discriminative visual
features for one of $T$ local image regions of a person bounding box image.
{\bf (2)} One {\em global branch}:
This aims to learn the optimal global level features from the entire person image.
For both branches, we select the Inception-A/B units \cite{xiao2016learning,szegedy2017inception} 
as the basic building blocks\footnote{This choice is independent of our model design
and others can be readily considered such as
AlexNet \cite{krizhevsky2012imagenet},
ResNet \cite{he2016deep} and VggNet \cite{simonyan2014very}.}.

In particular, we used 3 Inception-A and 3 Inception-B blocks for building the global branch, 
and 3 Inception-B blocks for each local stream.
The width (channel number) of each Inception is denoted by $d_1$, $d_2$ and $d_3$.
The global network ends with a \textit{global average pooling} layer and 
a \textit{fully-connected} (FC) feature layer with 512 outputs. 
For the local branch, we also use a 512-D FC feature layer 
which fuses the {global average pooling} outputs of all streams. 
To reduce the model parameter size, 
we share 
the first conv layer between global and local branches
and the same-layer Inceptions among all local streams.
For our HA-CNN model training, we utilise the {\em cross-entropy classification loss} function for both global and local branches, which optimise person identity classification.

For attention selection within each bounding box of some unknown misalignment,
we consider a {\em harmonious attention learning} scheme
that aims to jointly learn a set of complementary attention maps including
hard (regional) attention for the local branch and soft (spatial/pixel-level and channel/scale-level) attention
for the global branch.

We further introduce a {\em cross-attention interaction learning} scheme 
between the local and global branches
for further enhancing the harmony and compatibility degree
whilst
simultaneously optimising per-branch discriminative feature representations.

We shall now describe more details of each component of the network
design as follows.

\begin{figure} [!ht]
	\centering
	\includegraphics[width=0.8 \linewidth]{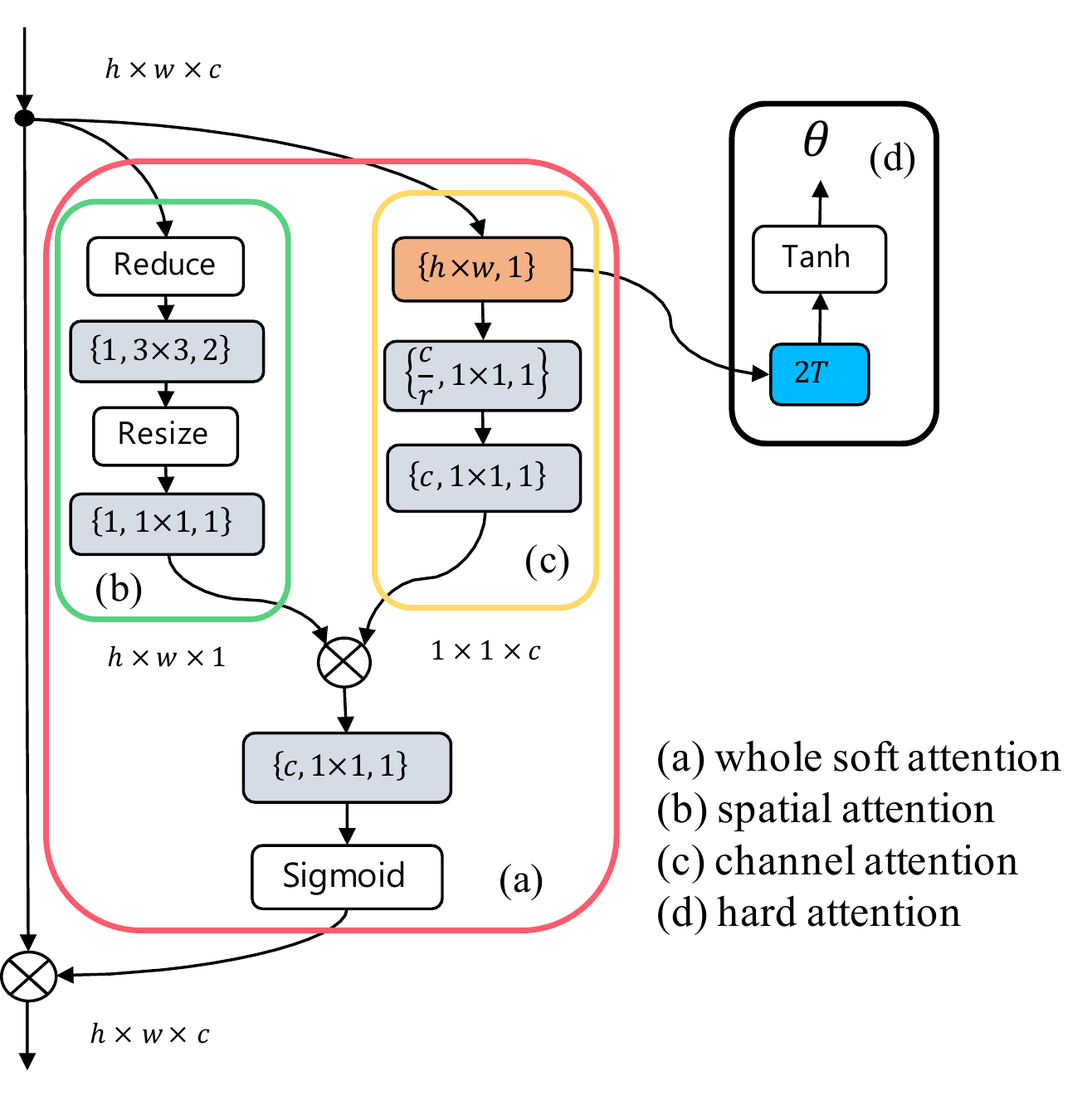}
	\vskip -0.3cm
	\caption{ %\footnotesize
		The structure of each Harmonious Attention module consists of
		{\bf (a)} Soft Attention which includes 
		{\bf (b)} Spatial Attention (pixel-wise)
		and 
		{\bf (c)} Channel Attention (scale-wise),
		and 
		{\bf (d)} Hard Regional Attention (part-wise).
		Layer type is indicated by background colour:
		\textcolor{battleshipgrey}{\bf grey} for \textit{convolutional} (conv),  
		\textcolor{brown}{\bf brown} for \textit{global average pooling}, 
		and 
		\textcolor{capri}{\bf blue} for \textit{fully-connected} layers.
		The three items in the bracket of a conv layer are:  
		filter number, filter shape, and stride.
		The ReLU \cite{krizhevsky2012imagenet} and Batch Normalisation (BN) \cite{ioffe2015batch} (applied to each conv layer) are not shown for brevity. 
	}
	\label{fig:attentions}
	\vspace{-0.5cm}
\end{figure}

\subsection{Harmonious Attention Learning}
\label{sec:method_attention}

Conceptually, our Harmonious Attention (HA) is a principled union
of {hard} {\em regional} attention \cite{jaderberg2015spatial},
soft {\em spatial} \cite{wang2017residual} and
{\em channel} attention \cite{hu2017squeeze}.
This simulates functionally the dorsal and ventral attention mechanism of human brain \cite{vossel2014dorsal} in the sense of modelling soft and hard attention simultaneously. 
The soft attention learning aims at selecting 
the {\em fine-grained} important pixels,
while the hard attention learning is dedicated 
to searching the {\em coarse} latent ({weakly supervised}) discriminative regions.
They are thus largely complementary
with high compatibility to each other in functionality.
Intuitively, their combination can relieve the modelling burden of soft attention
and resulting in more discriminative and robust model learning
from the same (particularly small) training data. 
We propose a novel {Harmonious Attention joint learning} strategy to unite 
the three distinct types of attention with only a small number of additional parameters.
We take a {\em block-wise} (module-wise) attention design,
that is, each HA module is specifically optimised to attend
the input feature representations at its own level alone.
In the CNN hierarchical framework, this naturally allows for 
{\em hierarchical} multi-level attention learning to progressively refine the attention maps, 
in the spirit of the divide and conquer design \cite{cormen2009introduction}.
As a result, we can significantly reduce the attention search space
(i.e. the model optimisation complexity) whilst allow
multi-scale selectiveness of hierarchical features to enrich the final feature representations.
Such progressive and holistic attention modelling is both intuitive and essential for person re-id 
due to that (1) the surveillance person images often have cluttered background and uncontrolled appearance variations therefore
the optimal attention patterns of different images can be highly variable,
and (2) a re-id model typically needs robust (generalisable) model learning given very
limited training data (significantly less than common image
classification tasks).
Next, we describe the design of our Harmonious Attention module in details.

\vspace{0.01cm}
\noindent {\bf (I) Soft Spatial-Channel Attention}
The input to a Harmonious Attention module is a 3-D tensor
$\bm{X}^{l} \in \mathcal{R}^{h \times w \times c}$
where $h$, $w$, and $c$ denote the number of pixel in
the height, width, and channel dimensions respectively;
and $l$ indicates the level of this module in the entire network
(multiple such modules).
Soft spatial-channel attention learning aims to produce a saliency weight map 
$\bm{A}^{l} \in \mathcal{R}^{h \times w \times c}$
of the same size as $\bm{X}$.
Given the largely independent nature between spatial (inter-pixel) and
channel (inter-scale) attention, 
we propose to learn them in a {\em joint} but {\em factorised} way as:
%, which allows us to epitomise the  advantages of both RAM and SE. 
\begin{equation}
\bm{A}^{l} = \bm{S}^{l} \times \bm{C}^{l}  
\end{equation}
where 
$\bm{S}^{l} \!\in\! \mathcal{R}^{h \times w \times 1}$ and
$\bm{C}^{l}\! \in \!\mathcal{R}^{1 \times 1 \times c}$ 
represent the spatial and channel attention maps,
% of the $l$-th level attention module, 
respectively. 
We perform the attention tensor factorisation 
by designing a two-branches unit (Fig. \ref{fig:attentions}(a)):
One branch to model the spatial attention $\bm{S}^{l}$ (shared across the channel dimension),
%for learning the pixel level importance in the height and width dimensions,
and another branch to model the channel attention $\bm{C}^{l}$ (shared across both height and width dimensions).

By this design, we can compute {\em efficiently} the full soft attention $\bm{A}^{l}$ 
from $\bm{C}^{l}$ and $\bm{S}^{l}$ 
with a tensor multiplication.
%
%Algorithmically, o
Our design is more efficient 
than common tensor factorisation algorithms
\cite{kolda2009tensor}
since heavy matrix operations are eliminated.
% and therefore much more computationally efficient.

\vspace{0.01cm}
\noindent {\bf \textit{(1) Spatial Attention}}
We model the spatial attention by a tiny (10 parameters) 4-layers sub-network
(Fig. \ref{fig:attentions}(b)).
It consists of a global cross-channel averaging pooling layer (0 parameter),
a conv layer of $3\times3$ filter with stride 2
(9 parameters),
%(for layer compression), 
a resizing bilinear layer (0 parameter),
%then reconstructed by bilinear \textit{resize}, 
and a scaling conv layer (1 parameter).
In particular, the global averaging pooling, defined as,
\begin{equation}
\bm{{S}_\text{input}}^{l} = \frac{1}{c} \sum_{i=1}^{c}  \bm{X}^{l}_{1:h, 1:w, i}
\end{equation}
is designed especially to compress the input size of the subsequent conv layer
with merely $\frac{1}{c}$ times of parameters needed.
This cross-channel pooling is reasonable because in our design all channels
share the identical spatial attention map.
We finally add the scaling layer for automatically learning an adaptive fusion scale 
in order to optimally combining the channel attention described next.

\vspace{0.01cm}
\noindent {\bf \textit{(2) Channel Attention}}
We model the channel attention by a small 
($2\frac{c^2}{r}$ parameters, see more details below) 4-layers
squeeze-and-excitation sub-network (Fig. \ref{fig:attentions}(c)).
%We adopt the squeeze-and-excitation design for modelling the channel attention \cite{hu2017squeeze}.
Specifically, we first perform a {\em squeeze} operation 
via an averaging pooling layer (0 parameters) for 
aggregating feature information distributed across the spatial space into
a channel signature as 
\begin{equation}
\bm{{C}_\text{input}}^{l} = \frac{1}{h \times w} \sum_{i=1}^{h}  \sum_{j=1}^{w} \bm{X}^{l}_{i,j,1:c}
%\in \mathcal{R}^{c \times 1}
\label{eq:squeeze}
\end{equation}
This signature conveys the per-channel filter response from the whole image,
therefore providing the complete information for the inter-channel dependency modelling
in the subsequent {\em excitation} operation, formulated as
\begin{equation}
\bm{{C}_\text{excitation}}^{l} =\texttt{ReLU}(\ \bm{W}_2^\text{ca} \times \texttt{ReLU}(\bm{W}_1^\text{ca} \bm{{C}_\text{input}}^{l}))
%\in \mathcal{R}^{c \times 1}
\end{equation}
where $\bm{W}_1^\text{ca} \in \mathcal{R}^{\frac{c}{r} \times c}$ 
($\frac{c^2}{r}$ parameters) and 
$\bm{W}_2^\text{ca} \in \mathcal{R}^{c \times \frac{c}{r}}$ 
($\frac{c^2}{r}$ parameters)
denote 
the parameter matrix of 2 conv layers in order respectively,
and $r$ ($16$ in our implementation) represents the bottleneck reduction rate.
Again, this bottleneck design is for reducing the model parameter
number from $c^2$ (using one conv layer) to $(\frac{c^2}{r}\!+\!\frac{c^2}{r})$,
e.g. only need $\frac{1}{8}$ times of parameters when $r=16$.

For facilitating the combination of the spatial attention and channel attention,
we further deploy  a $1\!\times \!1\!\times c$ convolution ($c^2$ parameters) layer to compute blended full soft attention after tensor multiplication. This is because the spatial and channel attention are not mutually exclusive but with a co-occurring complementary relationship. Finally, we use the sigmoid operation (0 parameter) to normalise the full soft attention into the range between 0.5 and 1.

\vspace{0.01cm}
\noindent {\bf \textit{Remarks}}
Our model is similar to 
the Residual Attention (RA) \cite{wang2017residual}
and Squeeze-and-Excitation (SE)  \cite{hu2017squeeze} concepts
but with a number of essential differences:
{\bf (1)} The RA requires to learn a much more complex soft attention sub-network which is 
not only computationally expensive but also less discriminative when
the training data size is small typical in person re-id.
{\bf (2)} The SE considers only the channel attention and implicitly
assumes non-cluttered background, therefore significantly restricting
its suitability to re-id tasks under
cluttered surveillance viewing conditions.
{\bf (3)} Both RA and SE consider no hard regional attention modelling, 
hence lacking the ability to discover the correlated complementary benefit
between soft and hard attention learning.

\vspace{0.01cm}
\noindent {\bf (II) Hard Regional Attention}
The hard attention learning
aims to locate latent ({\em weakly supervised}) discriminative
$T$ regions/parts (e.g. human body parts) in each input image at the $l$-th level.
We model this regional attention by learning a transformation matrix as:
\begin{equation}
\bm{A}^l = 
\begin{bmatrix}
s_h & 0 & t_x \\
0 & s_w & t_y \\
\end{bmatrix}
\label{eq:atheta}
\end{equation} 
which allows for image cropping, translation, and isotropic scaling operations
by varying two scale factors ($s_h$, $s_w$) and the 2-D spatial position ($t_x$, $t_y$).
We use pre-defined region size by fixing $s_h$ and $s_w$  for limiting the model complex.
Therefore, the effective modelling part of $\bm{A}^l$ is 
only $t_x$ and $t_y$, with the output dimension 
as $2\!\times\!T$ ($T$ the region number).
To perform this learning, we introduce
a simple 2-layers ($2\!\times\!T\!\times\!c$ parameters) sub-network
(Fig. \ref{fig:attentions}(d)).
We exploit the first layer output (a $c$-D vector) of the channel attention (Eq.~\eqref{eq:squeeze})
as the first FC layer ($2\!\times\!T\!\times\!c$ parameters) input for further reducing the parameter size 
while sharing the available knowledge
in spirit of the multi-task learning principle \cite{evgeniou2004regularized}.
The second layer ($0$ parameter) performs a \textit{tanh} scaling (the range of $[-1, 1]$)
to convert the region position parameters into the percentage
so as to allow for positioning individual regions outside of the input image boundary.
This specially takes into account the cases that 
only partial person is detected sometimes.
Note that, unlike the soft attention maps that 
are applied to the input feature representation $\bm{X}^l$,
the hard regional attention is enforced on that of the corresponding network block
to generate $T$ different parts which are subsequently fed into the corresponding 
streams of the {\em local} branch
(see the dashed arrow on the top of Fig \ref{fig:pipline}).

\begin{figure} %[!h]
	\vskip -0.3cm
	\centering
	\subfigure[\scriptsize STN \cite{jaderberg2015spatial}]{
		\includegraphics[width=0.5 \linewidth, height=0.20\linewidth]{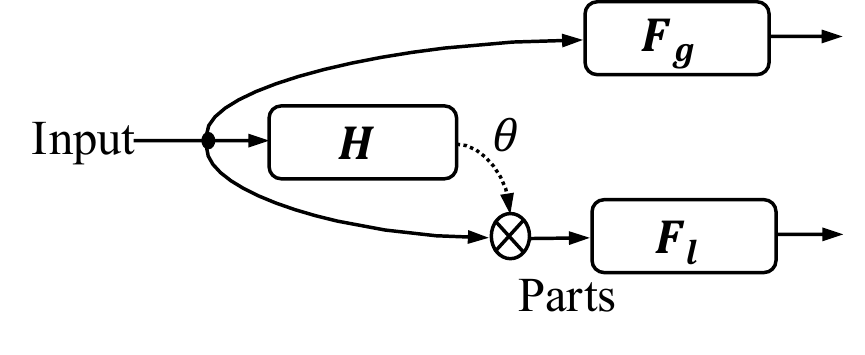}
		\label{fig:plain}
	}
	\subfigure[\scriptsize  HA-CNN Hard Attention]{
		\includegraphics[width=0.4 \linewidth, height=0.20\linewidth]{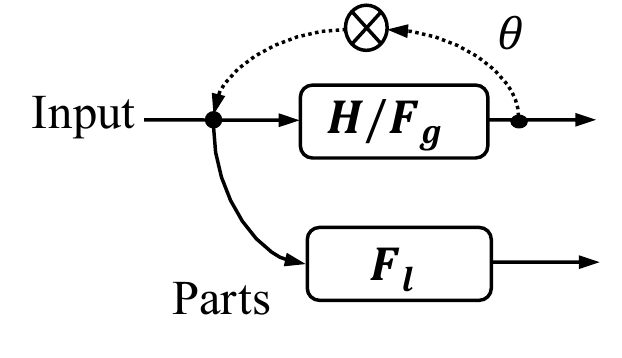}
		\label{fig:feedback}
	}
	\vskip -0.1cm
	\caption{ %\footnotesize
		Schematic comparison between (a) STN \cite{jaderberg2015spatial} 
		and (b) HA-CNN Hard Attention.
		Global feature and hard attention are jointly learned
                in a multi-task design.
		``$\bm{H}$'': Hard attention module;
		``$\bm{F_g}$'': Global feature module; 
		``$\bm{F_g}$'': Local feature module.
	}
	\label{fig:stns}
	\vspace{-0.5cm}
\end{figure}

\vspace{0.01cm}
\noindent {\bf \textit{Remarks}}
The proposed hard attention modelling is conceptually similar to
the Spatial Transformer Network (STN) \cite{jaderberg2015spatial}
because both are designed to learn a transformation matrix for discriminative region identification.
However, they differ significantly in design:
{\bf (1)} The STN attention is {\em network-wise} (one level of attention learning)
whilst our HA is {\em module-wise} (multiple levels of attention learning).
The latter not only eases the attention modelling complexity (divide-and-conquer design) and 
but also provides additional attention refinement in a sequential manner.
{\bf (2)} The STN utilises a separate large sub-network 
for attention modelling
whilst the HA-CNN exploits a much smaller sub-network 
by sharing the majority model learning with the target-task network
using a multi-task learning design (Fig. \ref{fig:stns}),
therefore superior in both higher efficiency and lower overfitting risk. 
{\bf (3)} The STN considers only hard attention 
whilst HA-CNN models both soft and hard attention in an end-to-end fashion
so that additional complementary benefits are exploited.

\vspace{0.01cm}
\noindent{\bf (III) Cross-Attention Interaction Learning}
Given the joint learning of soft and hard attention above, 
we further consider a cross-attention interaction mechanism
for enriching their joint learning harmony
by interacting the {\em attended} local and global features across branches.
Specifically, 
at the $l$-th level, 
we utilise the global-branch feature 
%$\bm{X}^{(l, k)}_{\text{global}}$
$\bm{X}^{(l, k)}_{G}$
of the $k$-th region to
enrich the corresponding 
local-branch feature
%$\bm{X}^{(l, k)}_{\text{local}}$
$\bm{X}^{(l, k)}_{L}$
by tensor addition as
%\begin{equation}
%\bm{\tilde{X}}^{(l, k)}_{\text{local}}  = \bm{X}^{(l, k)}_{\text{local}} + \bm{X}^{(l, k)}_{\text{global}}
%\end{equation}
\begin{equation}
\bm{\tilde{X}}^{(l, k)}_{L}  = \bm{X}^{(l, k)}_{L} + \bm{X}^{(l, k)}_{G}
\end{equation}
where $\bm{X}^{(l, k)}_{G}$ is computed by
applying the hard regional attention of the $(l\!+\!1)$-th level's HA attention module
(see the dashed arrow in Fig. \ref{fig:pipline}).
By doing so, we can simultaneously reduce the complexity of the local branch (fewer layers)
since the learning capability of the global branch can be partially shared.
During model training by back-propagation, 
the global branch takes gradients from both the global and local branches as

\begin{equation}\small
\Delta  \bm{W}^{(l)}_{G} = \frac{\partial \mathcal{L}_{G}}{ \partial \bm{X}^{(l)}_{G}} \frac{\partial \bm{X}^{(l)}_{G}}{ \partial \bm{W}^{(l)}_{G}} +  \sum_{k=1}^{T}  \frac{\partial \mathcal{L}_{L}}{ \partial \bm{\tilde{X}}^{(l, k)}_{L}}  \frac{\partial \bm{\tilde{X}}^{(l, k)}_{L}}{\partial \bm{W}^{(l)}_{G}} 
\end{equation}
Therefore, the global $\mathcal{L}_{G}$ and local $\mathcal{L}_{L}$ loss 
quantities are concurrently utilised in optimising the parameters $\bm{W}^{(l)}_{G}$
of the global branch.
As such, the learning of the global branch is interacted with that of
the local branch at multiple levels, whilst both are subject to the same re-id optimisation constraint.

\vspace{0.01cm}
\noindent {\bf \textit{Remarks}}
By design, cross-attention interaction learning is subsequent to and complementary
with the harmonious attention joint reasoning above. Specifically,
the latter learns soft and hard attention from
the same input feature representations to maximise their compatibility 
({\em joint attention generation}), whilst the former 
optimises the correlated complementary information between 
attention refined global and local features 
under the person re-id matching constraint
({\em joint attention application}).
Hence, the composition of both forms a complete process of 
joint optimisation of attention selection
for person re-id.
\subsection{Person Re-ID by HA-CNN}
Given a trained HA-CNN model,
we obtain a 1,024-D joint feature vector (deep feature representation)
by concatenating the local (512-D) and the global (512-D) feature vectors.
For person re-id, we deploy this 1,024-D deep feature representation using {\em only}
a generic distance metric {\em without} any camera-pair specific distance
metric learning, e.g. the L2 distance.
Specifically, given a test probe image $\bm{I}^p$ from one camera view
and a set of test gallery images 
$\{\bm{I}_i^g\}$ from other non-overlapping camera views:
(1) We first compute their corresponding 1,024-D feature vectors by
forward-feeding the images to a trained HA-CNN model, 
denoted as $\bm{x}^p=[\bm{x}_g^p; \bm{x}_l^p]$ and 
$\{\bm{x}_i^g=[\bm{x}_g^g; \bm{x}_l^g]\}$. 
(2) We then compute L2 normalisation on the global and local features,
respectively. 
(3) Lastly, we compute the cross-camera matching
distances between $\bm{x}^p$ and $\bm{x}_i^g$ by the L2 distance. 
We then rank all
gallery images in ascendant order by their L2 distances to the probe image.  
The probabilities of true matches of probe person images in Rank-1 and
among the higher ranks indicate the goodness of the learned
HA-CNN deep features for person re-id tasks.

%-------------------------------------------------------------------------
\section{Experiments}
\label{exp}
\vspace{-0.3cm}
\begin{figure} [ht]
	\centering
	\subfigure[CUHK03]{
		\includegraphics[width=0.24\linewidth]{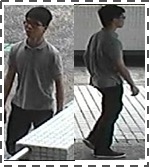}
	}
	\subfigure[Market-1501]{
		\includegraphics[width=0.24\linewidth]{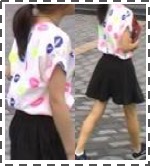}
	}
	\subfigure[DukeMTMC]{
		\includegraphics[width=0.24\linewidth]{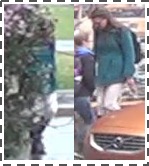}
	}
	\vskip -0.1cm
	\caption{ %\footnotesize
		Example cross-view matched pairs from three datasets.
	}
	\label{fig:dataset}
	\vspace{-0.3cm}
\end{figure}

%\subsection{Datasets,  Evaluation Protocols and Baselines}
\noindent {\bf Datasets and Evaluation Protocol}
For evaluation, % we evaluate our proposed HA-CNN model on 
we selected three large-scale person re-id benchmarks, Market-1501 \cite{wang2016highly}, DukeMTMC-ReID \cite{zheng2017unlabeled} and CUHK03
%\footnote{Instead of following the original split with 1,367/100 for training/testing for 20 rounds, we adopt the data split introduced in \cite{zheng2017unlabeled}.} 
\cite{li2014deepreid}.
% , which are widely-used to evaluate person re-identification. 
Figure \ref{fig:dataset} shows several example person bounding box images. % from these datasets. 
%
%\noindent {\bf Evaluation Protocols.} 
We adopted the standard person re-id setting
including the training/test ID split and test protocol
(Table \ref{tab:dataset_stats}).
For performance measure, we use
the cumulative matching characteristic (CMC) and mean Average Precision (mAP)
metrics.
\begin{table}[h!] %\footnotesize
	\centering
	%\scalebox{0.8}{
	\renewcommand{\arraystretch}{1.2}
	\setlength{\tabcolsep}{0.003cm}
	%\vspace{-0.5cm}
	\caption{%\footnotesize
		%Statistics and s
		Re-id evaluation protocol.
%		Settings of person re-id datasets.
		TS: Test Setting;
		SS: Single-Shot; MS: Multi-Shot.
		SQ: Single-Query; MQ: Multi-Query.
	}
	%\vskip 0pt %\vskip -6pt
	\begin{tabular}{|c||c|c|c|c||c|}
		\hline 
		Dataset  & 
%		{\# Cam} &
		{\# ID} & 
		{\# Train } & 
		{\# Test} &
		{\# Image} & 
		Test Setting \\ \hline \hline %
		%VIPeR & 2 & 632 & 316 & 316 & 1,264 & SS \\  % \hline
		%GRID & 8 & 250 & 125 & 125 & 1,275 & SS \\  % \hline
		%CUHK01 & 2 & 971 & 871/485 & 100/486 & 1,942& SS/MS\\
		CUHK03 %& 6 
		& 1,467 & 767 & 700 & 14,097  & SS\\
		Market-1501 %& 6  
		& 1,501& 751 & 750  & 32,668 & SQ/MQ \\
		DukeMCMT-ReID %& 8  
		& 1,402& 702 & 702 & 36,411 & SQ \\
		\hline
	\end{tabular}%}
	\label{tab:dataset_stats}
	\vspace{-0.3cm}
\end{table}

%\subsection{Implementation Details}
\noindent {\bf Implementation Details}
We implemented our HA-CNN model in the Tensorflow \cite{abadi2016tensorflow} framework. 
All person images are resized to $160\!\times\! 64$. 
For HA-CNN architecture, we set the width of Inception units 
at the $1^\text{st}$/$2^\text{nd}$/$3^\text{rd}$ levels as:
$d_1\!=\!128$,  $d_2\!=\!256$ and $d_3\!=\!384$.
Following \cite{li2017person}, we use $T\!=\!4$ regions for hard attention,
e.g. a total of $4$ local streams. In each stream, we fix the size of three levels of hard attention as $24\times28$, $12\times14$ and $6\times7$.
For model optimisation,
we use the ADAM \cite{kingma2014adam} algorithm  
at the initial learning rate $5\!\times\!10^{-4}$ with the two moment terms 
$\beta_1=0.9$ and $\beta_2=0.999$.
We set the batch size to 32,
epoch to 150,
momentum to 0.9.
Note, %for high training and test efficiency, 
we do {\bf \em not} adopt any data argumentation methods 
(e.g. scaling, rotation, flipping, and colour distortion),
{\bf \em neither} model pre-training.
Existing deep re-id methods typically benefit significantly from these operations
at the price of not only much higher computational cost but also
notoriously difficult and time-consuming model tuning.

\subsection{Comparisons to State-of-the-Art Methods}
%\subsubsection{Evaluation on Market-1501}
\noindent {\bf Evaluation on Market-1501}
We evaluated HA-CNN against 13 existing methods on Market-1501.
Table \ref{tab:res_market} shows the clear performance superiority of HA-CNN
over all state-of-the-arts with significant Rank-1 and mAP advantages. 
Specifically, HA-CNN outperforms the $2^\text{nd}$ best model JLML (pre-defined hard attention based)
% deep joint global-local CNN model (JLML) 
by $6.1\%$ (91.2-85.1) (SQ) and $4.1\%$ (93.8-89.7) (MQ) in Rank-1; $10.2\%$ (75.7-65.5) (SQ) and $8.3\%$ (82.8-74.5) (MQ) in mAP. 
Compared to the only soft attention alternative HPN,
our model improves the Rank-1 by $14.3\%$ (91.2-76.9) (SQ).
This indicates the superiority of our factorised spatial and channel soft attention modelling
over HPN's multi-directional attention mechanism.
HA-CNN also surpasses recent hard attention re-id methods (MSCAN,  DLPA and PDC),
boosting the Rank-1 by $10.9\%$, $10.2\%$ and $7.1\%$,
mAP by $18.2\%$, $12.3\%$ and $12.3\%$ (SQ), respectively. 
These validate the significant advantage of our harmonious soft/hand attention 
joint and interaction learning
% along with customised CNN architecture in an efficient end-to-end fashion,
over existing methods replying on either hard or soft attention at a single level. 
\vspace{-0.1cm}
\begin{table} %[!h]
	\centering
	%\footnotesize
	%\scalebox{0.8}{
	\renewcommand{\arraystretch}{1}
	\setlength{\tabcolsep}{0.28 cm} 
	%\vspace{-.5cm}
	\caption{%\footnotesize
		% Performance comparisons on 
		Market-1501 evaluation. $1^\text{st}/2^\text{nd}$ best in red/blue.
		% All person bounding box images were manually-cropped.
	}
	%\vskip 0pt %\vskip -6pt
	\begin{tabular}{|c|cc|cc|}
		\hline
		Query Type &  \multicolumn{2}{c|}{Single-Query} &\multicolumn{2}{c|}{Multi-Query} \\ \cline{1-5}
		Measure (\%)    & R1 & mAP & R1 & mAP  \\ \hline \hline
		XQDA\cite{liao2015person} &  43.8 &  22.2  &  54.1 &  28.4\\   %\hline
		SCS\cite{chen2016similarity} &  51.9 &  26.3 &  - &  -  \\  %\hline
		DNS\cite{zhang2016learning} & 61.0 & 35.6  &  71.5 &  46.0  \\ 
		CRAFT\cite{chen2017person} & 68.7 & 42.3  &  77.0 &  50.3  \\ 
		\hline
		CAN\cite{liu2017end} & 60.3 & 35.9  & 72.1 &  47.9  \\ 
		S-LSTM\cite{varior2016siamese} & - & -  & 61.6 &  35.3  \\ 
		G-SCNN\cite{varior2016gated} & 65.8 & 39.5  & 76.0 &  48.4  \\ 
		HPN \cite{liu2017hydraplus} & 76.9 & -  & - &  -  \\ 
		SVDNet \cite{sun2017svdnet}&  82.3 &  62.1 &- & -  \\ 
%		TriNet*~\cite{hermans2017defense} & 82.5 & 65.5 &- & - \\
%		TriNet$^{(10+)}$*~\cite{hermans2017defense} 
%		& 84.9 & \color{blue}69.1 & \color{blue}90.5 & \color{blue}76.4 \\
		\hline
		MSCAN \cite{li2017learning} & 80.3 & 57.5  &86.8 &  66.7  \\
		DLPA \cite{zhao2017deeply} & 81.0 & 63.4  & - &  -  \\ 
		PDC \cite{su2017pose} & 84.1 & 63.4  & - &  -  \\ 
		JLML \cite{li2017person} 
		& \color{blue}{85.1} &  \color{blue}{65.5}  & \color{blue} {89.7} & \color{blue}{74.5}  \\
%		IDEAL \cite{lan2017deep} & \color{blue} 86.7 & 67.5 & \color{blue} 91.3 & 76.2 \\ 
		\hline
%		{\bf HA-CNN-A} &  88.4&  70.4 &   92.5 &   78.3 \\ 
%		{\bf HA-CNN-B} &  90.9&  74.3 &   93.4 &   81.3 \\ 
		{\bf HA-CNN} & \color{red} {91.2} &  \color{red} {75.7}  &   \color{red} {93.8} &   \color{red} {82.8}  \\ 
		\hline
	\end{tabular}%}
	\label{tab:res_market}
	\vspace{-0.35cm}
\end{table}
%\subsubsection{
\vspace{0.05cm}

\noindent {\bf Evaluation on DukeMTMC-ReID}
We evaluated HA-CNN on the recently released DukeMTMC-ReID dataset\footnote{Only a small number of methods (see Table \ref{tab:res_duke_reid}) 
	have been evaluated and reported on DukeMTMC-ReID.}.
Compared to Market-1501, person images from this benchmark 
have more variations in resolution and background due to 
wider camera views and more complex scene layout,
therefore presenting a more challenging re-id task.
Table \ref{tab:res_duke_reid} shows that HA-CNN again outperforms
all compared state-of-the-arts with clear accuracy advantages,
surpassing the $2^\text{nd}$ best SVDNet-ResNet50 (without attention modelling) 
by $3.8\%$ (80.5-76.7) in Rank-1 
and $7.0\%$ (63.8-56.8) in mAP.
This suggests the importance of attention modelling in re-id 
and the efficacy of our attention joint learning approach in a
more challenging re-id scenario. 
Importantly, the performance advantage by our method is achieved 
at a lower model training and test cost through an much easier training process.
For example, the performance by SVDNet relies on 
the heavy ResNet50 CNN model (23.5 million parameters) 
with the need for model pre-training on the ImageNet data (1.2 million
images), whilst HA-CNN has only 2.7 million parameters with no data augmentation.

\begin{table} %[!h]
	\centering
	%\footnotesize
	%\scalebox{0.8}{
	\renewcommand{\arraystretch}{1}
	\setlength{\tabcolsep}{0.5cm}
	%\vspace{-.5cm}
	\caption{%\footnotesize
		% Performance comparisons on 
		DukeMTMC-ReID evaluation. $1^\text{st}/2^\text{nd}$ best in red/blue.
		%All person bounding box images were auto-detected.
	}
	%\vskip 0pt %\vskip -6pt
	\begin{tabular}{|c|cc|}
		\hline
		Measure (\%)    & R1 & mAP   \\ \hline \hline
		BoW+KISSME \cite{wang2016highly} &  25.1 &  12.2 \\   %\hline
		LOMO+XQDA \cite{liao2015person} &  30.8 &  17.0  \\ \hline
		ResNet50  \cite{he2016deep} &  65.2 &  45.0  \\ 
		ResNet50+LSRO \cite{zheng2017unlabeled}&  67.7 &  47.1  \\ 
		JLML \cite{li2017person} & 73.3 & 56.4   \\ 
		SVDNet-CaffeNet \cite{sun2017svdnet} & 67.6 & 45.8  \\
		SVDNet-ResNet50 \cite{sun2017svdnet} & \color{blue}{76.7} &   \color{blue}{56.8}  \\ 
		\hline
%		{\bf  HA-CNN-A} & 77.2 &  58.9 \\ 
%		{\bf HA-CNN-B} &  78.1 &  59.5  \\ 
		{\bf HA-CNN} &  \color{red} {80.5} &  \color{red} {63.8}  \\ 
		\hline
	\end{tabular}%}
	\label{tab:res_duke_reid}
	\vspace{-.3cm}
\end{table}

%\subsubsection{
\vspace{0.05cm}
\noindent {\bf Evaluation on CUHK03}
We evaluated HA-CNN on both manually labelled and 
auto-detected (more misalignment) person bounding boxes of the CUHK03 benchmark. 
We utilise the 767/700 identity split rather than 1367/100 since the former defines a more realistic and challenging re-id task.
In this setting, the training set is small with only about 7,300 images ({\em versus} 12,936/16,522 in Market-1501/DukeMCMT-ReID).
This generally imposes a harder challenge to deep models, particularly 
when our HA-CNN does not benefit from any auxiliary data pre-training
(e.g. ImageNet) nor data augmentation.
Table \ref{tab:res_cuhk03} shows that HA-CNN still achieves the best
re-id accuracy, outperforming hand-crafted feature based methods significantly
and deep competitors less so.
Our model achieves a small margin (+$0.2\%$ in Rank-1 and +$1.3\%$) over 
the best alternative SVDNet-ResNet50 on the detected set.
However, it is worth pointing out that SVDNet-ResNet50 benefits
additionally from not only large ImageNet pre-training but also a much
larger network and more complex training process. 
In contrast, HA-CNN is much more lightweight on parameter size with the advantage of easy training and fast deployment.   
This shows that our attention joint learning can be a better replacement of 
existing complex networks with time-consuming model training.

\begin{table} [!t]
	\centering
	%\footnotesize
	%\scalebox{0.8}{
	\renewcommand{\arraystretch}{1}
	\setlength{\tabcolsep}{0.2cm}
	%\vspace{-.5cm}
	\caption{%\footnotesize
		% Performance comparisons on 
		CUHK03 evaluation. The setting is 767/700 training/test split.
		$1^\text{st}/2^\text{nd}$ best in red/blue.
	}
	%\vskip 0pt %\vskip -6pt
	\begin{tabular}{|c|cc|cc|}
		\hline
		\multirow{2}{*}{Measure (\%)}
		&  \multicolumn{2}{c|}{Labelled} &\multicolumn{2}{c|}{Detected} \\ \cline{2-5}
		
		& R1 & mAP & R1 & mAP  \\ \hline \hline
		BoW+XQDA \cite{wang2016highly} &  7.9 &  7.3 &  6.4 &  6.4\\   %\hline
		LOMO+XQDA \cite{liao2015person} &  14.8 &  13.6 &  12.8 &  11.5\\   \hline
		IDE-C \cite{zhong2017re} &  15.6 &  14.9 &  15.1 &  14.2\\   %\hline
		IDE-C+XQDA \cite{zhong2017re}&  21.9 &  20.0 &  21.1 &  19.0\\   %\hline
		IDE-R \cite{zhong2017re}& 22.2 &  21.0 &  21.3 &19.7 \\   %\hline
		IDE-R+XQDA \cite{zhong2017re} &  \color{blue}32.0 & \color{blue} 29.6 &  31.1 &  28.2\\   %\hline
		SVDNet-CaffeNet \cite{sun2017svdnet} &  - &  -  &  27.7  &  24.9 \\ 
		SVDNet-ResNet50 \cite{sun2017svdnet} &  - &  -  &  \color{blue} {41.5} &  \color{blue} {37.3} \\ 
		\hline
%		HA-CNN-B  &  - & - & - &  -  \\ 
%		HA-CNN-B  &  43.5 & 40.1 & 39.8 &   36.8  \\ 
		{\bf HA-CNN} &  \color{red} {44.4} &  \color{red} {41.0}  & \color{red}{41.7} &   \color{red} {38.6}  \\ 
		\hline
	\end{tabular}%}
	\label{tab:res_cuhk03}
	%\vspace{-.3cm}
\end{table}

%-------------------------------------------------------------------------
\subsection{Further Analysis and Discussions} 
\label{sec:fad}

\noindent {\bf Effect of Different Types of Attention}
We further evaluated the effect of each individual attention component
in our HA model:
Soft Spatial Attention (SSA), Soft Channel Attention (SCA), and Hard Regional Attention (HRA).
Table \ref{tab:res_am} shows that: 
{\bf (1)} Any of the three attention {\em in isolation} brings person re-id performance gain;
{\bf (2)} The combination of SSA and SCA gives further accuracy boost, which suggests
the complementary information between the two soft attention discovered
by our model;
{\bf (3)} When combining the hard and soft attention, another significant performance gain
is obtained. This shows that our method is effective in identifying
and exploiting the complementary information between 
coarse hard attention and fine-grained
soft attention.

\begin{table} % [!h]
	\centering
	%\footnotesize
	%\scalebox{0.8}{
	\renewcommand{\arraystretch}{1}
	\setlength{\tabcolsep}{0.30 cm}
	%\vspace{-.5cm}
	\caption{%\footnotesize
		%Performance comparisons of different attention modules on Market-1501 and DukeMTMC-ReID in 
		Evaluating individual types of attention in our HA model.
		{\em Setting}: SQ.
%		Base: the basic global model without any attention modules.
		SSA: Soft Spatial Attention;
		SCA: Soft Channel Attention;
		HRA: Hard Regional Attention.
	}
	\vskip -1pt %\vskip -6pt
	\begin{tabular}{|c|cc|cc|}
		\hline
		Dataset &  \multicolumn{2}{c|}{Market-1501} &
		\multicolumn{2}{c|}{DukeMTMC-ReID} \\ \cline{1-5}
		Metric (\%)    & R1 & mAP & R1 & mAP  \\ \hline \hline
		No Attention &  84.7 &  65.3  &  72.4 &  53.4\\  \hline
		SSA & 85.5 & 65.8 & 73.9  & 54.8 \\
		%RAM \cite{wang2017residual}  &  86.1 &  68.2  &  75.4 &  56.7\\ %\hline
		%
		SCA & 86.8 & 67.9  &  73.7 &  53.5 \\
		SSA+SCA  &  88.5 &  70.2  &  76.1 &  57.2 \\ \hline
		%CA \cite{hu2017squeeze} & 86.8 & 67.9  &  73.7 &  53.5  \\ 
		HRA & 88.2 & 71.0 & 75.3 & 58.4
		\\
		\hline
		\bf All & \bf 91.2 & \bf 75.7  & \bf 80.5 & \bf 63.8\\\hline
		
%		Base+STN (G) &  86.4/90.8 &  67.6/76.3  &   &  \\ %\hline
%		Base+STN (L) &  86.2/90.7 &  66.5/75.1  &   &  \\ %\hline
%		Base+STN (G+L) &  88.2/91.6 &  71.0/78.8  &   &  \\ %\hline
%		\hline
	\end{tabular}%}
	\label{tab:res_am}
	\vspace{-0.3cm}
\end{table}

%\subsubsection{Effects of Hierarchical Interactions}
\vspace{0.02cm}
\noindent {\bf Effect of Cross-Attention Interaction Learning}
We also evaluated the benefit of cross-attention interaction learning (CAIL) between global and local branches. 
Table \ref{tab:res_hi} shows that CAIL has significant benefit to re-id matching,
improving the Rank-1 by $4.6\%$(91.2-86.6) / $6.5\%$(80.5-74.0),
mAP by $9.5\%$(75.7-66.2) / $8.4\%$(63.8-55.4) on Market-1501 / DukeMTMC-ReID, respectively.
This validates our design is rational that it is necessary to jointly learn
the {\em attended} feature representations across soft and hard attention
subject to the same re-id label constraint.
%We conclude that hierarchical interactions enhance the power of cooperative learning inside HA-CNN.
\vspace{-0.1cm}
\begin{table} [!t]
	\centering
	%\footnotesize
	%\scalebox{0.8}{
	\renewcommand{\arraystretch}{1.2}
	\setlength{\tabcolsep}{0.2 cm}
	%\vspace{-.5cm}
	\caption{%\footnotesize
		Evaluating cross-attention interaction learning (CAIL). {\em Setting}: SQ.
	}
	\vskip 1pt %\vskip -6pt
	\begin{tabular}{|c|cc|cc|}
		\hline
		Dataset &  \multicolumn{2}{c|}{Market-1501} &
		\multicolumn{2}{c|}{DukeMTMC-ReID} \\ \cline{1-5}
		Metric (\%)    & R1 & mAP & R1 & mAP  \\ \hline \hline
%		Base &  84.7 &  65.3  &  72.4 &  53.4\\  
%		Base+FSAM &  88.5 &  70.2  &  76.1 &  57.2\\ \hline
		%
		{\bf w/o} CAIL &  86.6 &  66.2  &  74.0 &  55.4\\  
		\hline
		{\bf w/} CAIL & \bf 91.2 & \bf 75.7  & \bf 80.5 & \bf 63.8\\ 
		\hline
	\end{tabular}%}
	\label{tab:res_hi}
	\vspace{-0.4cm}
\end{table}

\vspace{0.01cm}
\noindent {\bf Effect of Joint Local and Global Features}
We evaluated the effect of joint local and global features by
comparing their individual re-id performances against that of the joint feature.
Table \ref{tab:G_L} shows: 
{\bf (1)} Either feature representation {\em alone} is already very
discriminative for person re-id. For instance, the global HA-CNN feature 
outperforms the best alternative JLML \cite{li2017person}
(Table \ref{tab:res_market}) by $4.8\%$(89.9-85.1) in Rank-1 
and by $7.0\%$(72.5-65.5) in mAP (SQ) on Market-1501.
{\bf (2)} A further performance gain is obtained by joining the two
representations, yielding $6.1\%$(91.2-85.1) in Rank-1 boost and $10.2\%$(75.7-65.5) in mAP increase.
Similar trends are observed on the DukeMCMT-ReID (Table \ref{tab:res_duke_reid}).
These validate the complementary effect of jointly learning
local and global features in harmonious attention context 
by our HA-CNN model. 

\vspace{-0.3cm}
\begin{table} [!h]
	\centering
	%\footnotesize
	%\scalebox{0.8}{
	\renewcommand{\arraystretch}{1.2}
	\setlength{\tabcolsep}{0.2 cm}
	%\vspace{-.5cm}
	\caption{%\footnotesize
		Evaluating global-level and local-level features. {\em Setting}: SQ.
	}
	\vskip 1pt %\vskip -6pt
	\begin{tabular}{|c|cc|cc|}
		\hline
		Dataset &  \multicolumn{2}{c|}{Market-1501} &\multicolumn{2}{c|}{DukeMTMC-ReID} \\ \cline{1-5}
		Metric (\%)    & R1 & mAP & R1 & mAP  \\ \hline \hline
		Global &  89.9 &  72.5  &  78.9 &  60.0\\ %\hline
		Local &  88.9 &  71.7  &  77.3 &  59.5\\ %\hline
		\bf Global+Local & \bf 91.2 & \bf 75.7  & \bf 80.5 & \bf 63.8\\ 
		\hline
	\end{tabular}%}
	\label{tab:G_L}
	\vspace{-0.3cm}
\end{table}

\begin{figure}[!h]
	\centering
	\vspace{-.3cm}
	\includegraphics[width=0.68 \linewidth]{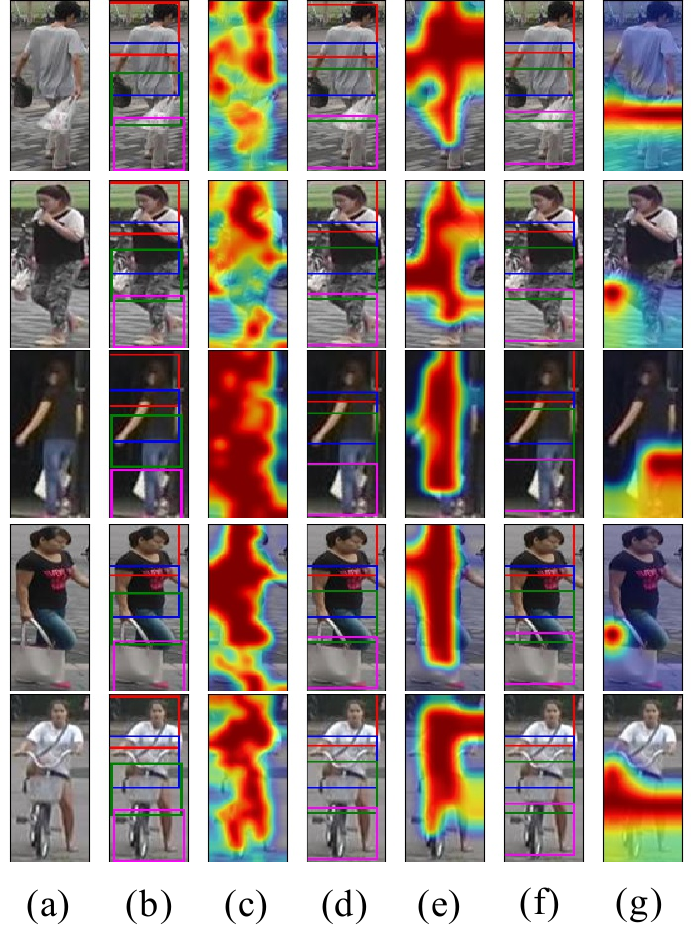}
	\vspace{-.3cm}
	\caption{Visualisation of our harmonious attention in person re-id.
		From left to right, (a) the original image, (b) the $1^{\text{st}}$-level
                of HA, (c) the $1^{\text{st}}$-level of SA,
                (d) the $2^{\text{nd}}$-level of HA, 
		(e) the $2^{\text{nd}}$-level of SA,  (f) the $3^{\text{rd}}$level of
                  HA, (g) the $3^{\text{rd}}$-level of SA.
	}
	\label{fig:visa}
	\vspace{0.02cm}
\end{figure}

%\subsubsection{
\vspace{-0.2cm}
\noindent {\bf Visualisation of Harmonious Attention} We visualise both learned soft attention and hard attention at three different levels of HA-CNN. Figure \ref{fig:visa} shows:  {\bf (1)} 
Hard attention localises four body parts well at all three levels,
approximately corresponding to head+shoulder (\textcolor{red}{red}), upper-body (\textcolor{blue}{blue}), upper-leg (\textcolor{green}{green})
and lower-leg (\textcolor{violet}{violet}). {\bf (2)} Soft attention focuses on the discriminative pixel-wise selections progressively in spatial localisation, e.g. attending hierarchically from the global whole body by the $1^{\text{st}}$-level spatial SA (c) to local salient parts (e.g. object associations) by the $3^{\text{rd}}$-level spatial SA (g). This shows compellingly the effectiveness of joint soft and hard attention learning.
\vspace{-0.3cm}

%\subsubsection{
\vspace{0.2cm}
\noindent {\bf Model Complexity}
We compare the proposed HA-CNN model with 
four popular CNN architectures
(Alexnet \cite{krizhevsky2012imagenet},
VGG16 \cite{simonyan2014very},
GoogLeNet \cite{szegedy2015going}, and
ResNet50 \cite{he2016deep}) in model size and complexity.
Table \ref{tab:base_nets} shows that 
HA-CNN has the smallest model size (2.7 million parameters) 
and the $2^\text{nd}$ smallest FLOPs ($1.09\!\times\! 10^9$)
and yet, still retains the $2^\text{nd}$ deepest structure (39).

\begin{table} [!h]
	\centering
	% \scriptsize
	%\footnotesize
	%\scalebox{0.8}{
	\renewcommand{\arraystretch}{1}
	\setlength{\tabcolsep}{0.2 cm}
	\vspace{-0.3cm}
	\caption{ %\footnotesize
		Comparisons of model size and complexity.
		FLOPs: the number of FLoating-point OPerations;
		PN: Parameter Number.
	}
	%\vskip 0pt %\vskip -6pt
	\begin{tabular}{|c||c|c|c|c|}
		\hline
		Model  & FLOPs & PN (million) & Depth 
		\\ \hline
		AlexNet \cite{krizhevsky2012imagenet}
		& \bf 7.25$\times$10$^8$ & 58.3 & 7  \\ % & 239.8\\   %\hline
		VGG16 \cite{simonyan2014very}
		&1.55$\times$10$^{10}$ & 134.2 & 16  \\ % & 549.4 \\   %\hline
		ResNet50 \cite{he2016deep}
		& 3.80$\times$10$^9$ & 23.5 & \bf 50 \\ % & 100.4 \\   
		GoogLeNet \cite{szegedy2015going}
		& 1.57$\times$10$^9$ & 6.0 & 22  \\  % & 50.5 \\ 
		JLML &1.54$\times$10$^9$ & 7.2 & 39 \\ 
		\hline
%		HA-CNN-A &\bf4.54$\times$10$^8$ &  \bf 1.0 & 38 \\ 
%		HA-CNN-B &8.10$\times$10$^8$ &1.6 & 38 \\
		\bf HA-CNN &1.09$\times$10$^9$ & \bf 2.7 & 39 \\ % & \bf 34.6\\ 
		\hline
	\end{tabular}%}
	\label{tab:base_nets}
	\vspace{-0.35cm}
\end{table}

%-------------------------------------------------------------------------
\section{Conclusion}
In this work, we presented a novel Harmonious Attention Convolutional Neural Network (HA-CNN)
for joint learning of person re-identification attention selection and feature representations
in an end-to-end fashion.
In contrast to most existing re-id methods
that either ignore the matching misalignment problem or 
exploit stringent attention learning algorithms, 
the proposed model is capable of extracting/exploiting multiple
complementary attention and maximising their latent complementary effect 
for person re-id in a unified {\em lightweight} CNN architecture.
This is made possible by the Harmonious Attention module design
in combination with a two-branches CNN architecture.
Moreover, we introduce a cross-attention interaction learning mechanism
to further optimise joint attention selection and re-id feature
learning.
Extensive evaluations were conducted on three re-id benchmarks to validate the advantages of
the proposed HA-CNN model over a wide range of state-of-the-art methods 
on both manually labelled and more challenging auto-detected person images.
We also provided detailed model component analysis and
discussed HA-CNN's model complexity as compared to popular alternatives.

{\small
\bibliographystyle{ieee}
\bibliography{reid}
}

\end{document}